# Application of Causal Inference to Analytical Customer Relationship Management in Banking and Insurance

Satyam Kumar [a,b] and Vadlamani Ravi [a,*]

[a] Center for AI and ML,
Institute for Development and Research in Banking Technology (IDRBT), Castle Hills,
Masab Tank, Hyderabad 500057, India
[b] School of Computer and Information Sciences (SCIS), University of Hyderabad,
Hyderabad 500046, India
learnsatyam@gmail.com;vravi@idrbt.ac.in

## Abstract

Of late, in order to have better acceptability among various domain, researchers have argued that machine intelligence algorithms must be able to provide explanations that humans can understand causally. This aspect, also known as 'causability' achieves a specific level of human-level explainability. A specific class of algorithms known as counterfactuals may be able to provide causability. In statistics, causality has been studied and applied for many years, but not in great detail in artificial intelligence (AI). In a first-of-its-kind study, we employed the principles of causal inference to provide explainability for solving the analytical customer relationship management (ACRM) problems. In the context of banking and insurance, current research on interpretability tries to address causality-related questions like why did this model make such decisions, and was the model's choice influenced by a particular factor? We propose a solution in the form of an intervention, wherein the effect of changing the distribution of features of ACRM datasets is studied on the target feature. Subsequently, a set of counterfactuals is also obtained that may be furnished to any customer who demands an explanation of the decision taken by the bank/insurance company. Except for the credit card churn prediction dataset, good quality counterfactuals were generated for the loan default, insurance fraud detection, and credit card fraud detection datasets, where changes in no more than three features are observed.

*Keywords:* Causal Inference; Analytical Customer Relationship Management; Classification; Counterfactuals

## 1. Introduction

In many areas of human activity, such as science, engineering, commerce, and law, causal information is highly valued and desired. Compared to other animals, homo sapiens' ability to learn, interpret, and use causal information is possibly one of the defining characteristics of human intelligence (Penn &

---
* Corresponding Author: Tel.: +914023294310; Fax: +914023535157

Povinelli, 2007). In his most recent book, Pearl made a persuasive case for the importance of causal reasoning, saying, for example (Pearl & Mackenzie, 2018, p1): "Some tens of thousands of years ago, people began to realize that certain things influence other things and that tinkering with the former can change the latter... This finding led to the development of structured civilizations, followed by the development of towns and cities, and ultimately the science and technology-based civilization we enjoy today."

On an intuitive level, the ability to analyze causal information is fundamental to human cognition from early development on, playing a crucial role in higher-level cognition, allowing us to organize an action plan, assign blame, identify who is to blame and who is responsible, and generalize across changing contexts. More specifically, it enables us to comprehend ourselves, interact with others, and make sense of the environment (Bareinboim et al., 2022).

The study of computers that can learn and think like people is the goal of artificial intelligence (AI), which is perhaps the oldest and most comprehensive branch of computer science. It's frequently referred to as "machine intelligence" (Poole et al., 1998), to distinguish it from "human intelligence" (Russell et al., 2009). This field focuses on the intersection of cognitive science and computer science (Tenenbaum et al., 2011). AI is gaining popularity as a result of practical breakthroughs in machine learning (ML).

Machine learning models remain black boxes (Ribeiro et al., 2016) and are incompetent to explain the reasons for their predictions or recommendations. Actual cause-and-effect relationships are important in artificial intelligence applications. The automatic generation of adequate explanations, an essential task in natural language planning, diagnosis, and processing, thus requires a formal analysis of the real-cause concept (Halpern & Pearl, 2000).

Regression, estimation, and hypothesis testing approaches serve as examples of standard statistical analysis, whose objective is to assess a distribution's parameters from samples taken from it. These characteristics can be used to infer relationships between variables, estimate beliefs or probability of past and future events, and update those probabilities in light of new evidence or measurements. As long as the experimental settings don't vary, the traditional statistical analysis does a great job of managing these tasks (Pearl, 2020). Statistics cannot be used to infer a cause unless both the factor of interest (X) and the result (Y) are measurable (Holland, 1986). The goal of causal analysis is to infer not only beliefs or probabilities under static conditions but also the dynamics of beliefs under changing conditions, such as changes made by treatments or by external interventions. In general, the laws of probability theory do not describe how one feature of distribution ought to change when another attribute is modified. A distribution function doesn't inform us how the distribution would change if the external conditions were changed, such as changing from an observational to an experimental setup. Causal hypotheses must identify links that hold up in the face of changing external conditions to give this information (Pearl, 2020).

## 1.1. Transition from Explainability to Causality

Instead of focusing on causation, traditional interpretable machine learning focuses on associations. As causal inference has grown in popularity, interpretable machine learning has seen an increase in the number of causality-oriented algorithms that have been suggested. Comparatively to traditional methods, causal approaches can be used to determine the causes and effects of model design or to undertake an analysis of its choices and actions (Xu et al., 2020).

A model must provide the decision-maker with an explanation and ensure that those explanations accurately show the true reasons for the model's decision to be considered interpretable (Serrano & Smith, 2019). Current XAI models (also known as post hoc, model-agnostic models) seek to understand a black box that has already been trained by constructing machine learning models based on local interpretations, providing approximations to the predictive black box (Ribeiro et al., 2016), (Lundberg & Lee, 2017), rather than representing the genuine underlying mechanisms of the black box (Rudin, 2019). To approximate the predictions of the black box, these algorithms compute correlations between specific features. The primary distinction between XAI and Causal-based AI is that the former seeks only to explain how a model might arrive at a prediction by weighting the available features, but the latter seeks to uncover the process governing the system under investigation to provide insight.

A common theory of causation, known as Granger causality (Granger, 1969) has been utilized extensively to infer causal relationships from time series data (Hiemstra & Jones, 1994; Ding et al., 2006). The theory behind Granger causality is that Y Granger causes X if it holds some unique information about X that is not present in X's past.

This paper will primarily focus on the causal effect because it is more strongly related to machine learning interpretability. In this study, a structural network produced by CausalNex for the dataset of banking and insurance will focus on a change that will be reflected in the distribution of the target feature by modifying the distribution of a feature. Visualization of the structure network will help the banking and insurance-based companies to know which features is most likely to cause an effect on another feature. Banking and insurance-based businesses generally provide a single counterfactual example to the user. For the provided instance, DiCE will generate a set of counterfactual which is lacking in the CausalNex tool.

The rest of the paper is structured as follows: The literature review study is covered in Section 2. The background knowledge necessary to understand concepts is discussed in Section 3. The data description is presented in section 4. The experimental setup is described in Section 5. Results for each data are presented in Section 6. The conclusion is presented in Section 6.

## 2.　　Literature survey

Bayesian networks have the potential to be used in fundamental financial analysis. and it was applied to a particular sector of the Czech economy engineering. In this work, (Gemela, 2003). The use of Bayesian networks on the risk and return of a portfolio has shown that finance models place a heavy focus on the quantitative, historical correlations between economic elements (Shenoy & Shenoy, 2003).In the context of systemic risk, inference and application of Bayesian graphical models are discussed. Here, emphasis was placed on the connections and potential applications of network models in financial econometrics (Ahelegbey, 2016).

The research is based on the application of the canonical correlation approach for structuring causal relationships between the indicators for the assessment of the stability of the banking system, which are divided into four sub-indices (assessing the intensity of credit and financial interaction in the interbank market, the efficacy of the banking system functions, structural changes and financial disproportions in the banking system, and activities of systemically important institutions) (Kolodiziev et al., 2018) .

The underlying ML model is assumed to be stationary in most counterfactual explanation generating methods. This isn't always the case, though, as credit card firms and banks often change their machine learning (ML) models (Garisch,2019), and this needs to be considered when giving someone counterfactuals.

An unique optimization formulation that, to explain the predictions of the black-box model, by creating sparse counterfactual explanations using a customized genetic algorithm and using this method can not only be applied to explain both approved and failed loan applications, not just those that were denied (Dastile, 2022).

Both the pre-merger market equilibrium and the post-merger environment, where this model was used to develop counterfactuals, are taken into consideration while estimating a credit market structural model. It looked at the two loan market segments for households and companies individually (Barros et al., 2014).

According to (Keane & Smyth, 2020), an endogenous method for creating counterfactuals can be created by altering native counterfactuals from the dataset. A good counterfactual was defined as having at most 2 desirable features.

# 3. Overview of Proposed Methodology

## 3.1. Pearl's hierarchy of Causal Models

Judea Pearl, a computer scientist, developed the Ladder of Causation, a framework that focuses on distinct roles of seeing, doing, and imagining, which led to a breakthrough in causality understanding. This is termed as Pearl Causal Hierarchy (PCH) and it has three hierarchical levels of causal models namely Association, Intervention and Counterfactuals. Different kinds of questions can be answered at each level, and it is required to have a foundational understanding of the lower levels to respond to questions at the upper levels (Pearl, 2019, p 54). In reality, we would expect to be able to respond to intervention and association-type questions before being able to answer retrospective queries. The focus of layer 1 of PCH is based on observational and factual information. In terms of effects of actions, Layer 2 encodes information about what might occur, if any intervention were to take place. Finally, Layer 3 answers hypothetical questions regarding what would have happened if a certain intervention had been made if something else had happened (Bareinboim et al., 2022, p.5).

## 3.2. Bayesian Network

Bayesian Network is a kind of probabilistic model, which can handle both discrete and continuous variables. A Bayesian network comprises nodes which represent variables and edges that indicate the probability relationship between the different elements. Causal Networks were later developed as a result of Bayesian Belief Networks (BBN). There are cases where it is viewed as a causal network. To progress from the Association to the Intervention level in the Causality Hierarchy, Bayesian Belief Networks are one of the primary Machine Learning approaches that can be applied. Connections between conditionally dependent and independent variables can be expressed using BBN. The frequentist method would be reached if an infinite amount of data were available since, in Bayesian statistics, the weight of our prior belief gradually diminishes as more data is presented. However, when discussing causality analysis, this case is not valid.

Currently, there is a lot of research focused on using Bayesian Belief Networks as a starting point to develop Causal Bayesian Networks (CBN) by companies notably (DeepMind, 2019). To visually detect and numerically quantify unfairness trends in datasets, CBN is employed.

### 3.3. Counterfactuals

Other disciplines including philosophy, psychology, and the social sciences have a long history of using counterfactual explanations. In 1973, philosophers like David Lewis wrote articles on the concept of counterfactuals (Parry & Lewis, 1973). According to psychological research (Byrne, 2007; Byrne, 2019; Kahneman & Miller, 2002). Counterfactuals cause people to think about causal relationships. In addition, philosophers have used counterfactuals to support the idea of causal thinking (Demopoulos, 1982; Woodward, 2004).

The example-based model-agnostic method is provided through a counterfactual explanation. In this kind of explanation, new instances are generated that contradict the models' prediction. The goal of counterfactual explanation provided as an optimization problem (Wachter et al., 2018), where it maximizes the following function from the set of points P to produce the set of counterfactual samples.

$$\mathcal{L}(x, x', y', \lambda) = arg \min_{x} \max_{\lambda} \left( \lambda \cdot \left( \hat{f}(x') - y' \right)^2 + d(x_i, x') \right) \tag{1}$$

$$d(x_i, x') = \sum_{k \in F} \frac{|x_k - x'_k|}{MAD_k}$$

$$MAD_k = median_{j \in P} |x_{j,k} - median_{l \in P}(x_{l,k})| \tag{2}$$

Where d (. ) is the distance function which is used for calculating Manhattan distance weighted between the instance x to be explained and the counterfactual x' and this distance function is scaled by the inverse of the median absolute deviation (MAD). $MAD_k$ defined for feature k, over the set of points P is given in equation 2. In equation 1, first half pushes the counterfactual of x away from the original prediction while the second half keeps the counterfactual close to the original prediction.

If a feature k varies drastically throughout the dataset, a synthetic point x' may also vary this feature while still being close to xi according to the distance metric. This metric is more resistant to outliers since median absolute difference is used instead of standard deviation, which is more commonly used.

## 4.     Description of datasets

This section describes briefly the classification datasets namely Loan default, Churn Prediction, Credit card fraud detection, and Insurance fraud.

### 4.1.   Loan Default dataset

From April 2005 to September 2005, the payment information was obtained from a bank located in Taiwan (a cash and credit card provider) with the bank's credit card holders as the intended audience. Out of a total of 30000 records, 6636 records (or 22.12%) represent cardholders that have missed payments. The target variable is a binary variable called default payment (Yes = 1, No = 0) (Yeh & Lien, 2009).

## 4.2. Credit card Churn Prediction dataset

A Latin American bank that has seen a rise in customer revenue and has been working on improving its retention strategy. This data, which was collected from the University of Chile organized Business Intelligence Cup (*Business Intelligence Cup, 2004*), has 14814 instances, of which 13812 instances are loyal customers, while the remaining 1002 are churners (disloyal consumers), accounting for 93% and 7%, respectively, of the total number of customers.

## 4.3. Insurance Fraud detection dataset

The Automobile insurance fraud dataset (Pyle, 1999) is freely available, powered by Angoss Knowledge Seeker software. This dataset is highly unbalanced, with 94% belonging to non-fraudulent customers and 6% belonging to fraudulent customers. This dataset was collected in 2 phases. Data collected in phase 1 was for the period January 1994 to December 1995, totaling 11338 records; while in phase 2 data were collected from January 1996 to December 1996, totaling 4083 records. There are 6 numeric and 25 categorical features present in the original dataset, including the target variable Fraud Found (1= Fraud, 0=Non-fraud).

## 4.4. Credit card fraud detection dataset

The dataset contains transactions made by European credit cardholders in September 2013 over 2 days (Pozzolo et al., 2014). The dataset contains records of 284,807 transactions, of which 492 were fraudulent. This dataset is highly unbalanced, with 0.172% of all transactions being fraudulent. This dataset has features $V_1$, $V_2$… and $V_{28}$ as principal components obtained by principal component analysis (PCA). PCA does not apply to Time and Amount. The time between each transaction and the first transaction in the data set is measured in seconds. The target variable, class, is mapped to the following: 1 belongs to fraudulent transactions and 0 belongs to non-fraudulent transactions.

## 5. Experiments

Fig.1 describes the methodology followed to carry out the experiments on four classification datasets. Experiments have been performed using two causal inference tools namely CausalNex and Diverse Counterfactual Explanation (DiCE). The following hyperparameters were set in CausalNex: proximity was set to 0.3 and diversity was set to 3.2 for each of the datasets.

## 5.1. CausalNex

Assuming constant/fixed relationship effects, the bulk of existing causal approaches frequently have trouble taking into account the complicated relationships between variables. By enabling data scientists to create models that take into account how changes in one variable may affect other variables using Bayesian Networks. CausalNex is a new Python package that addresses this difficulty (Beaumont et al., 2021). This not only enables us to identify plausible links between variables but also demonstrates these relationships through causal models. Directed acyclic graphs (DAGs) serve as a visual representation of a causal model that shows the link between variables. CausalNex creates a DAG using optimized structure learning algorithm, NOTEARS (Zheng et al., 2018), through which the complexity reduces from exponential to O ($d^3$), where d is the number of nodes in DAG.

## 5.2. Diverse Counterfactual

To aid a person in understanding a complex machine learning model, we need a set of counterfactual examples for any example-based decision support system (Kim et.al, 2016). These examples should, in theory, strike a balance between a wide variety of suggested changes (diversity), the relative ease of accepting those changes (proximity to the original input), and the causal laws of human society, such as the difficulty of changing one's race or educational level.

A trained machine learning model (In our case, random forest), f, and an instance, x, are the inputs. In order to produce a set of k counterfactual examples $\{c_1, c_2, ..., c_k\}$ , where each one leads to a different conclusion than x, it is required to construct a number of counterfactual examples. The instance (x) and each of the CF examples $\{c_1, c_2, ..., c_k\}$ are of d-dimensional (Mothilal et al., 2020).

The chance that at least one of a diverse group of CF examples will be useable by the user improves; examples may change a substantial number of features or aim for maximum diversity by taking important input considerations into account. It employs the determinantal point processes (DPP) based metric that has already been used to solve subset selection problem (Kulesza & Taskar, 2012). It is represented as follows: $dpp_{diversity} = \det(K)$, where $\det(.)$ is the determinant of kernel matrix (K) for the given counterfactuals. Kernel matrix is represented as follows: $K_{i,j} = \frac{1}{1+dist(c_i,c_j)}$, where dist ($c_i$, $c_j$) is the distance between two counterfactuals examples $c_i$ and $c_j$.

The most helpful CF examples for a user may be those that are most similar to the original input. Proximity is the negative vector distance between the original input and CF example's features. It is represented as $Proximity := \frac{-1}{k} \sum_{i=1}^{k} dist(c_i, x)$, where k is the set of counterfactual examples.

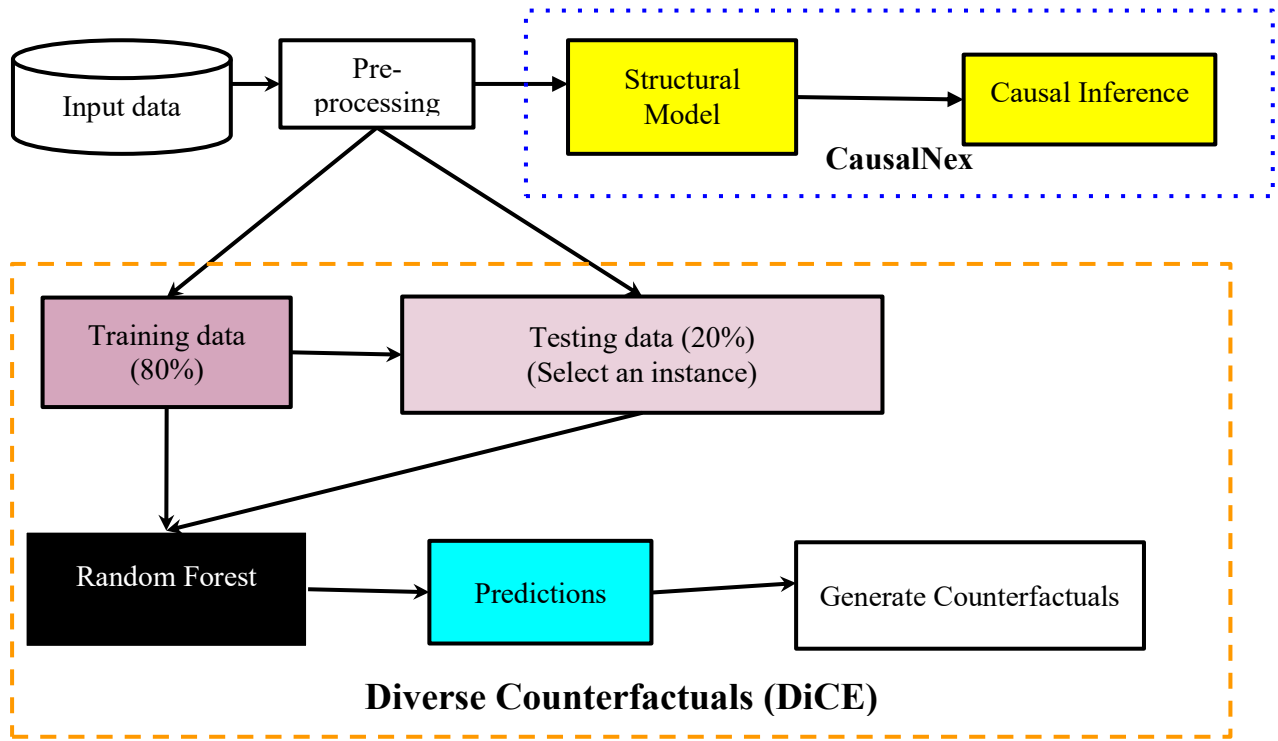

Fig. 1. Schematic representation of using CausalNex and DiCE.

## 6.    Results and Discussion

The symbol for do-intervention is P (Y|do(X)), which means if X were to be set at any value at random, what would be the probability distribution of Y? Do-intervention is incredibly helpful in counterfactual analysis, when we are curious to know if the results would have changed if we had taken an alternative action or intervention.

Some features in the data of loan defaults have even more categories than those listed in the data description. The feature EDUCATION, which contains categorical values of 0, 5, and 6, is mapped to the value 4 because contrary to the data description, the supplied feature has six categories in the dataset. As 0 and -2 values are not provided in the data description, features like PAY_0, PAY_1, PAY_2, PAY_3, PAY_4, PAY_5, and PAY_6 which have categorical values of 0 and -2 were mapped to value -1.

In the dataset for insurance fraud, it can be seen that the age feature reappears twice, once in numerical form (Age) and once in categorical form (Age of Policyholder), with the numeric values removed from the data to reduce the complexity possessed by multiple unique values (Farquad et al., 2012). Policy Number, Month, Week of Month, Day of Week, Day of Week Claimed, Week of Month Claimed, and Year are among the features that have been removed from the dataset. Features like Time were excluded from the dataset used to detect credit card fraud.

## 6.1. Loan Default dataset

The name of the features of this dataset to be used for Causal inference has been renamed as follows: i → j, where i is the original feature name, while j is the renamed feature. It is applicable only for the CausalNex tool, as long feature names are not permissible in that tool.

LIMIT_BAL →LB, PAY_i, where $i \in$ {0,2,3,4,5,6} → Pi, BILL_AMTj, →BAi, PAY_AMTj, → PAi, where j ∈ {1,2, , 6} default.payment.next.month →isdefaulter.

CausalNex creates structure from the data for the features having categorical values, Therefore, the numeric features were converted into discrete form for categorizing based on the mean value of these features, which is presented in Table 1.

Fig.2 depicts that one of the DAG obtained by using CausalNex is selected, where we wanted to check how applying intervention on any one of the features affects the distribution of target (or outcome). It shows that PA1, PA2, PA3, PA4, PA5, PA6, BA1, BA4, BA5, BA6, AGE, MARRIAGE, EDUCATION and SEX are the causes which are likely to cause an effect on the target variable (isdefaulter). This DAG is not ideal because, without the feature BA2 and BA3, PA2 and PA3 cannot be used to determine changes in the target variable.

Table 2 presents the intervention for the features LB, MARRIAGE, PA4, and BA5. Target variables are distributed so that 55.7% of them belong to non-defaulters and the remaining 44.3% are in the default class. There is a possibility of a decrease in loan defaulter from 44.3% to 44% if the distribution of LB<167484.323 is adjusted from the original 57 % distribution to 20 %. This suggests that a slight change in the LIMIT BAL's value would have little effect on the distribution of defaulters. Decrease in defaulter from 44.3 % to 42.3 %t is shown if the distribution of singles is altered from 53% to 50%, married people are changed from 45% to 30%, and others are increased from 2% to 20%.

Table 1. Mapping of numeric features into binary values based on the mean values for Loan default.

| Feature Name | Renamed Feature | Mapping |
|---|---|---|
| PAY_AMT1 | PA1 | **0:** PA1<5663.580, **1:** 5663.580<=PA1< 873552.1 |
| PAY_AMT2 | PA2 | **0:** PA2<5921.163", **1:** 5921.163<=PA2<1684259.1 |
| PAY_AMT3 | PA3 | **0:** PA3<5225.681, **1:** 5225.681<=PA3<896040.1 |
| PAY_AMT4 | PA4 | **0:** PA4<4826.076, **1:** 4826.076<=PA4<621000.1 |
| PAY_AMT5 | PA5 | **0:** PA5<4799.387, **1:** 4799.387<=PA5<426529.1 |
| PAY_AMT6 | PA6 | **0:** PA6<5215.502, **1:** 5215.502<=PA6<528666.1 |
| BILL_AMT1 | BA1 | **0:** BA1< 51223.330, **1:** 51223.330<=BA1<964511.1 |
| BILL_AMT2 | BA2 | **0:** BA2<49179.075, **1:** 49179.075<=BA2<983931.1 |
| BILL_AMT3 | BA3 | **0:** BA3< 47013.15, **1:** 47013.15<=BA3<1664089.1 |
| BILL_AMT4 | BA4 | **0:** BA4<43262.948, **1:** 43262.948<=BA4<891586.1 |
| BILL_AMT4 | BA5 | **0:** BA5<40311.400, **1:** 40311.400<=BA5<927171.1 |
| BILL_AMT6 | BA6 | **0:** BA6<38871.760, **1:** 38871.760<=BA6<961664.1 |
| AGE | AGE | 0: AGE<36, 1: 36<AGE<80 |
| LIMIT_BAL | LB | **0:** LB<167484.323, **1:** 167484.323<=LB<1000000.1 |

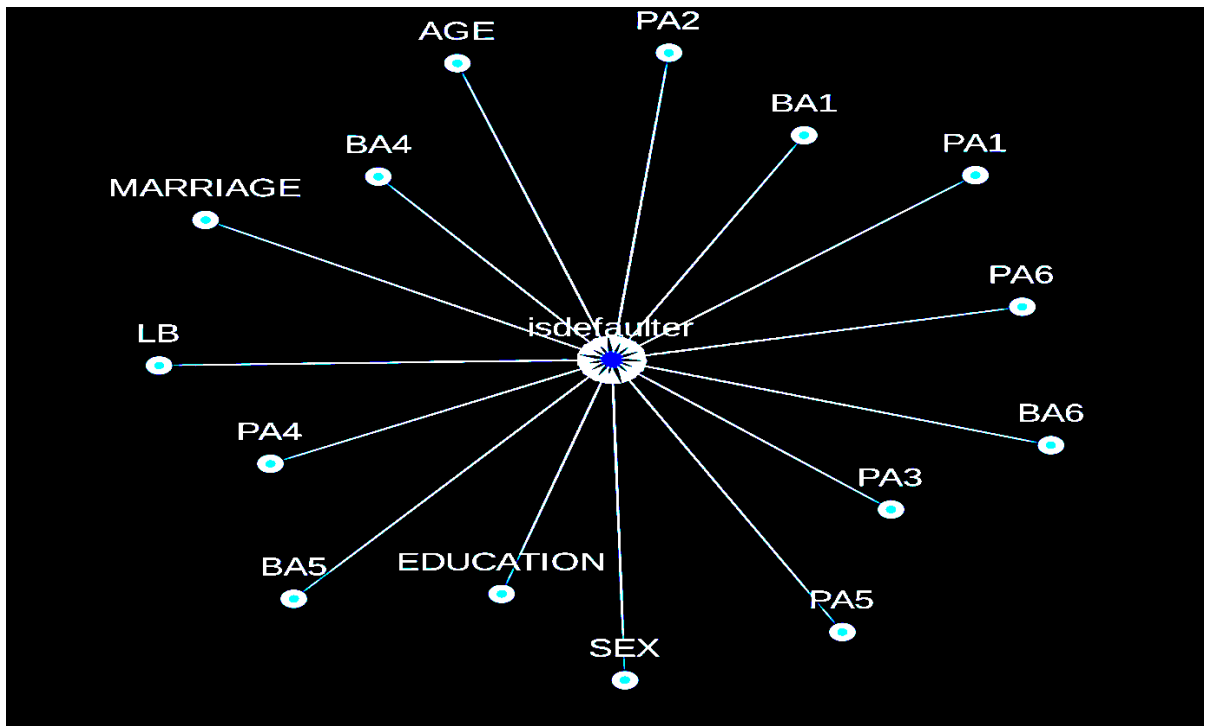

Fig. 2. DAG obtained using CausalNex for Loan default dataset

Table 2. Applied Interventions on features of Loan default dataset.

| Features | Category | Intervention | Changed p.d* of target |
|---|---|---|---|
| LB | 0: LB<167484.323, 1: 167484.323<=LB<1000000.1 | Original distribution of LB: 0:0.57,1:0.43 Changed distribution of LB: 0: 0.2, 1:0.8 | 0: 0.56 , 1: 0.44 |
| MARRIAGE | 0=Married, 1=Single, 2=others | Original distribution of MARRIAGE: 0: 0.45, 1: 0.53, 2: 0.02 Changed distribution of feature: 0: 0.3,1: 0.5, 2:0.2 | 0: 0.576, 1: 0.423 |

*p.d =Probability distribution

Following assumptions are made while generating the counterfactuals for this dataset

- If counterfactuals generated for the given instance on the feature AGE is less than that in original dataset ( $X_{AGE}^{CF} \leq X_{AGE}$ ), then it should be ignored as AGE for the selected instance cannot decrease and thus generate invalid counterfactuals.

- If counterfactuals generated for the given instance based on the category of EDUCATION feature is less than that in original dataset ( $X_{EDUCATION}^{CF} \leq X_{EDUCATION}$ ), then it should be ignored as EDUCATION for the selected instance cannot decrease and lead to invalid counterfactuals.

- Sex generated by the counterfactuals should also be the same as the selected instance.

Table 3 presents that for the 2nd test instance, the target variable is non-defaulter using Random forest as the black box model.. 4 counterfactuals generated using DiCE. Highlighted rows are the counterfactual values of the features. It has been observed that changing the category of PAY_4 feature and also changing the value of other features such as LIMIT_BAL, BILL_AMT2, and BILL_AMT5 is more likely to change the value of target from non-defaulter to defaulter. Remaining features values are kept unchanged.

Table 3. Counterfactuals generated by DiCE for Loan default dataset, where CFi is the counterfactual.

| Features | Original Values | CF 1 | CF2 | CF3 | CF4 |
|---|---|---|---|---|---|
| LIMIT_BAL | 150000 | | | 302091.6 | |
| PAY_4 | 0 | 7 | 4 | 8 | 1 |
| PAY_5 | 0 | | | | |
| PAY_6 | 0 | | | | |
| BILL_AMT1 | 15000 | | | | |
| BILL_AMT2 | 0 | 888474.3 | | | |
| BILL_AMT3 | 0 | | | | |
| BILL_AMT4 | 0 | | | | |
| BILL_AMT5 | 11694 | | 41966 | | 887023 |
| Target | 0 | 1 | 1 | 1 | 1 |

## 6.2.    Credit card Churn Prediction dataset

The name of the features of this dataset to be used for Causal inference has been renamed as follows: All the features having _ and - in the feature were removed and resultant feature name is the renamed feature. The numeric features were converted into discrete forms for categorizing based on the mean value of these features. The numeric features were converted into discrete forms for categorizing based on the mean value of these features, which is presented in Table 4.

Table 4. Mapping of numeric features based on the mean values for Churn Prediction.

| Feature Name | Renamed Feature | Mapping |
|---|---|---|
| INCOME | INCOME | 0:INCOME<889.772715,1:889.772715<=INCOME<10403 |
| AGE | AGE | **0:** AGE<36.491, **1:** 36.491<=AGE<77.1 |
| T_WEB_T-1 | TWEBT1 | 0:TWEBT1<6.259147 1: 6.259147<=TWEBT1<2491 |
| T_WEB_T-2 | TWEBT2 | 0:TWEBT2<5.831983, 1:5.8319831<=TWEBT2<191.1 |
| CRED_T | CREDT | 0: CREDT<631.652, 1: 0.877<=CREDT<12.1 |
| CRED_T-1 | CREDT1 | 0:CREDT1<129.490460,1:129.490460<=CREDT1<4359.6 |
| CRED_T-2 | CREDT2 | 0:CREDT2<128.470986,1:128.470986<=CREDT2<6814.3 |
| NCC_T | NCCT | 0:  NCCT<0.881 , 1:  0.881<=NCCT1<1.1 |

| NCC_T-1 | NCCT1 | 0: NCCT1<0.877, 1: 0.877<=NCCT<12.1 |
| NCC_T-2 | NCCT2 | 0: NCCT2<0.88, 1: 0.883<=NCCT2<12.1 |
| MAR_T | MART | 0: MART<13.63932, 1: 13.639327<=MART<5549.39 |
| MAR_T-1 | MART1 | 0: MART1<6.903797, 1: 6.903797<=MART1<2919.731 |
| MAR_T-2 | MART2 | 0:MART2<11.61460, 1: 11.614606<=MART2<3557.331 |
| MAR_T-3 | MART3 | 0: MART3<10.052, 1: 10.052<=MART3<2193.93 |
| MAR_T-4 | MART4 | 0: MART4<5.83, 1: 5.83<=MART4<2297.1 |
| MAR_T-5 | MART5 | 0: MART5<8.767 , 1: 8.767<=MART5<2394.16 |
| MAR_T-6 | MART6 | 0:MART6<12.390898,1: 12.390898<=MART6<2002.081 |

Fig.3 depicts that the DAG obtained by using CausalNex, where applying intervention on any one of the features affects the distribution of target (or outcome) is checked. It shows that features such as SX, INCOME, NCCT, NCCT1, NCCT2, MART, MART1, MART2, MART3, MART4, MART5, MART6, INCOME, CREDT, and CREDT2 are some of the causes which are likely to cause an effect on the target variable (Target).

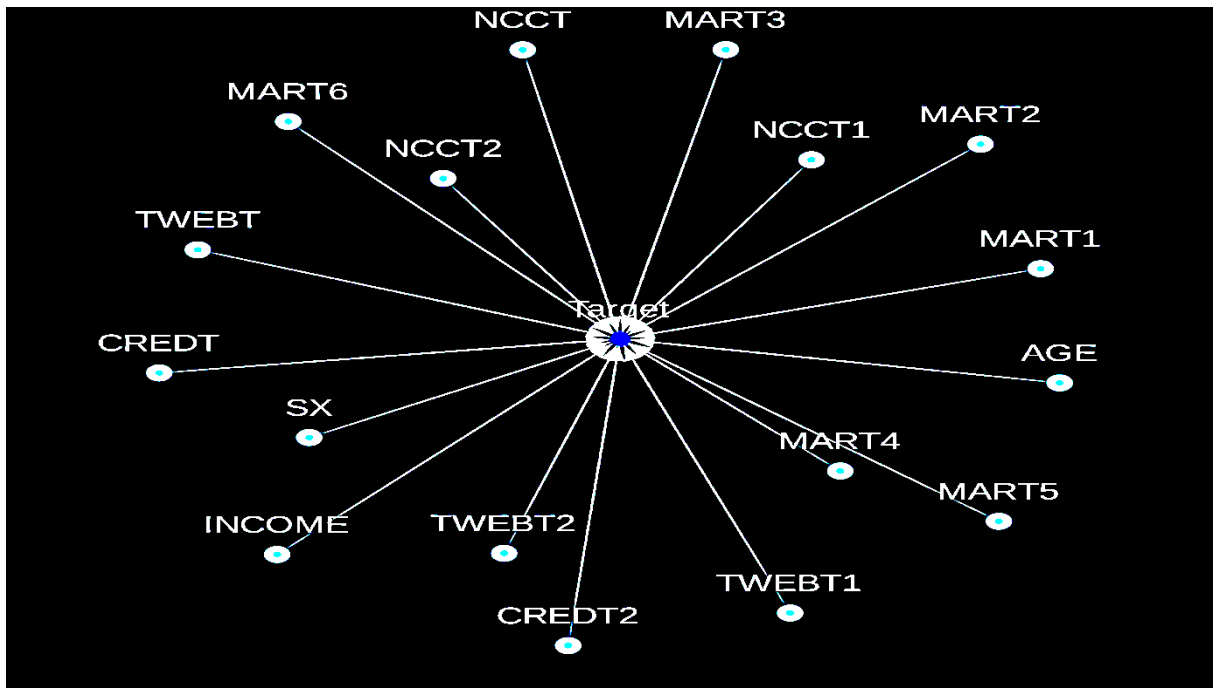

Fig. 3. DAG obtained using CausalNex for Churn Prediction dataset.

Table 5 presents the intervention for the features MART6, AGE. Target variables are distributed so that 93.2% of them belong to non-churners (loyal customers) and the remaining 6.8% are in the disloyal customer. There is a possibility of an increase in churners from 6.8% to 44.6% if the distribution of either of MART6<12.391 is adjusted from the original 63.6 % distribution to 40 % or 36.491<AGE < 77.1 is changed from 40.9% to 60%. This suggests that older people are more likely to churn than younger, changing the customer margin in the sixth month will result in more churners than at present. Company needs to ensure that proper margin for customers is set so that there are fewer churners possible.

Table 5. Intervention on features of Churn prediction dataset.

| Features | Category | Intervention | Changed p.d* of target |
|----------|----------|--------------|------------------------|
| MART6 | 0: MART6<12.391, 1:12.391<=MART6<2002.081 | Original distribution of MART6:0: 0.636,1:0.364 Changed distribution of MART6 :0: 0.4, 1:0.6 | 0: 0.446, 1: 0.554 |
| AGE | 0: AGE<36.491, 1: 36.491<=AGE<77.1 | Original distribution of AGE: 0:0.591,1:0.409 Changed distribution of AGE : 0: 0.4, 1:0.6 | 0: 0.44, 1: 0.56 |

*p.d = Probability distribution.

Following assumptions are made while generating the counterfactuals for this dataset

• If counterfactuals generated for the given instance on the feature AGE is less than that in original dataset ( $X_{AGE}^{CF} \leq X_{AGE}$ ), then it should be ignored as AGE for the selected instance cannot decrease and generate invalid counterfactuals.

• If counterfactuals generated for the given instance based on the category of N_EDUC feature is less than that in original dataset ( $X_{NEDUC}^{CF} \leq X_{NEDUC}$ ), then it should be ignored as N_EDUC for the selected instance cannot decrease and lead to invalid counterfactuals.

• SX generated by the counterfactuals should also be the same as the selected instance.

• For counterfactuals to be valid following changes are feasible E_CIV status can change from single to married (from 1 to 2), married to divorce (from 2 to 4) or married to widow (2 to 3). Remaining changes lead to invalid counterfactuals.

Table 6 presents that for the 1st test instance, the target variable is non-defaulter using Random forest as the black box model. 4 counterfactuals generated using DiCE. Highlighted rows present the counterfactual values of the features. Changing values of features such as TWEBT, TWEBT1 and TWEBT2 are few of the causes, which is more likely to change target from churner to non-churner. Counterfactual 2 (CF2) is invalid, as the category of EDUCATION for the given instance has changed from 3 to 4 and is represented by orange color column.

Table 6. Counterfactuals generated by DiCE for Churn Prediction, where CFi is the counterfactual.

| Features | Original Values | CF 1 | CF2 | CF3 | CF4 |
|---|---|---|---|---|---|
| CREDT | 601.80 | | | | |
| CREDT1 | 125.81 | 3353.4 | | 3353.4 | |
| CREDT2 | 95.44 | | 1169.67 | | |
| NCCT | 0 | | | | |
| NCCT1 | 0 | | | | |
| NCCT2 | 0 | 1 | | | |
| INCOME | 750.00 | | 9757.5 | | |
| EDU | 3.00 | | 4 | | |
| AGE | 30.00 | | 60.8 | 74.4 | |
| SX | 1.00 | | | | |
| ECIV | 1.00 | 2 | | 2 | |
| TWEBT | 10.00 | 104 | 32 | 104 | 104 |
| TWEBT1 | 0 | 24.1 | 203.9 | 24.1 | 24.1 |
| TWEBT2 | 2.00 | 107.5 | 18.7 | 107.5 | 107.5 |
| MART | -752.48 | | 4213 | | |
| MART1 | -16.13 | 2121.6 | -2289.2 | 2121.6 | |
| MART2 | 27.61 | | -2576.2 | | |
| MART3 | 34.02 | | -145.3 | | |
| MART4 | 18.51 | | | -6838.7 | |
| MART5 | 26.34 | -1323.3 | | -1323.3 | |
| MART6 | 31.08 | | -274.2 | | |
| Target | 1 | 0 | 0 | 0 | 0 |

## 6.3.  Insurance Fraud detection dataset

The name of the features of this dataset to be used for Causa has been renamed as follows:  i → j , where i is the original feature name, while j is the renamed feature. It is applicable only for the CausalNex tool, as long feature names are not permissible in that tool.

AccidentArea→AccArea,   MaritalStatus→MS,   PolicyType→PT,   VehicleCategory→VC, Days:Policy-Accident→DPA,   Days:Policy-Claim→DPC,   PastNumberOfClaims→PastNC, AgeOfVehicle→VehAge,   AgeOfPolicyHolder→APH,   PoliceReportFiled→PRF, WitnessPresent→WP,   AgentType→AT,   NumberOfSuppliments→NS,   AddressChange-Claim→ACC,   NumberOfCars→Ncars,   BasePolicy→BP,   FraudFound→isFraud, VehiclePrice→VPrice, RepNumber→RepNo, Deductible→Ded,  DriverRating→DR

The numeric features were converted into discrete forms for categorizing based on the mean value of these features. Following shows the mapping done for all the numeric features. The numeric features were converted into discrete forms for categorizing based on the mean value of these features, which is presented in Table 7.

Table 7. Mapping of numeric features into binary values based on the mean values for Insurance fraud.

| Feature Name | Renamed Feature | Mapping |
|---|---|---|
| AccidentArea- | AccArea | 0: AccArea< 0.8956, 1: 0.8956<=AccArea<1.1 |
| Make | Make | 0:  Make<10.111 , 1:  10.111<=Make<18.1 |
| BasePolicy | BP | 0:  BP<1.037 , 1:  1.037<=BP<2.1 |
| RepNumber | RepNo | 0:  RepNo<8 , 1:  8<=RepNo<16.1 |
| Deductible | Ded | 0:  Ded<407.644293 , 1:  407.644293<=Ded< 700.1 |
| PolicyType, | PT | 0:  PT<0.003605 , 1:  0.003605<=PT< 6.1 |
| VehiclePrice | VPrice | 0:  VPrice<0.003431 , 1:  0.003431<=VPrice< 3.1 |

Fig. 4 depicts that one of the DAG obtained by using CausalNex is selected, where we wanted to check how applying intervention on any one of the features affects the distribution of target (or isFraud). It shows that PT, Sex, MS, BP, DPC, WP, AccArea, VPrice, VehAge, Make, AT,are the causes which are likely to cause an effect on the target variable (isFraud). Also DPA is the cause of VPrice. Without a police complaint, there is no need for a witness, but in this case a witness is present during the accident. Accident type should also be effected by presence of witness, which is missed in this DAG.

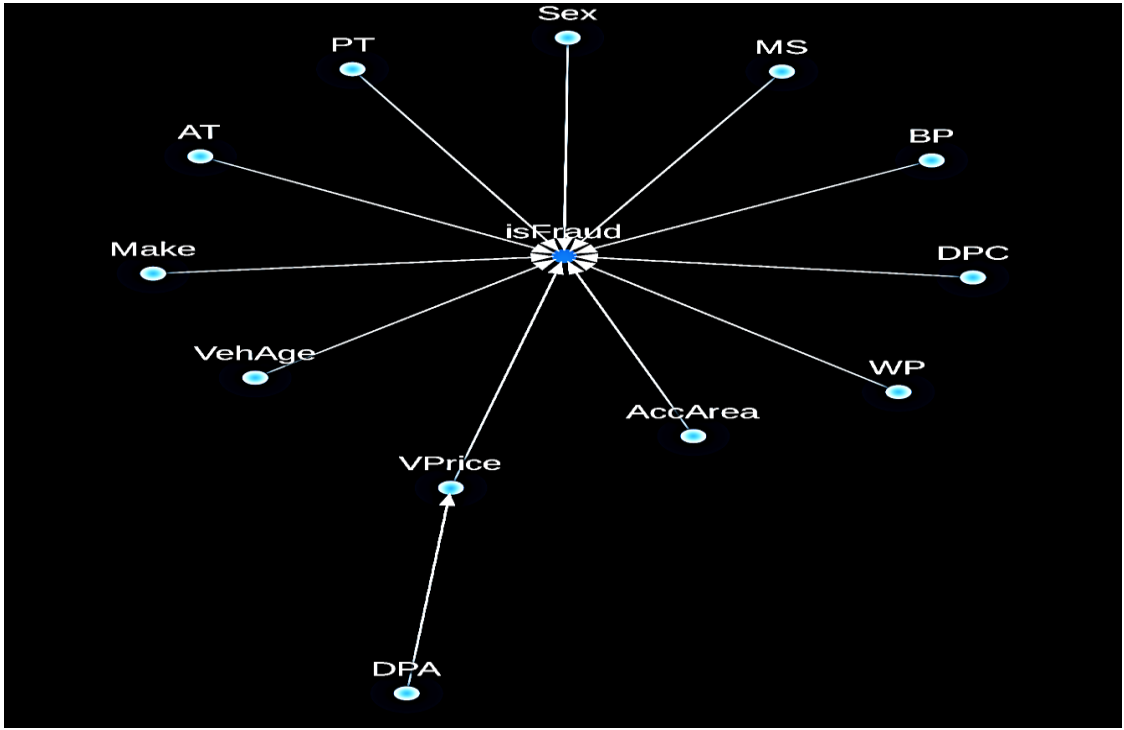

Fig. 4. DAG obtained using CausalNex for Auto Insurance fraud dataset.

Table 8 presents the intervention for the features Make, SEX. With respect to these 2 features, Target variables (isFraud) are distributed so that 64.1% of them belong to non-fraudulent insurance and the remaining 35.9% are fraudulent insurance. There is a possibility of an increase in fraudulent insurance from 35.9% to 43.9% if the distribution of 10<Make<18.1 is adjusted from 4.2% in the original distribution to 85%. Also, changing the distribution of Male is decreased from 91% to 70%, will increase the number of fraudulent insurance from 35.9% to 36.8%. This indicates that the issue of fraudulent insurance is going to become more prevalent in car types including Mercedes, Mercury, Nissan, Pontiac, Porsche, Saab, Saturn, Toyota, and VW. Therefore, firms should exercise caution while providing insurance for these car models.

Table 8. Interventions on features of Insurance Fraud dataset. (p.d = Probability distribution)

| Features | Category | Intervention | Changed p.d* of target |
|----------|----------|--------------|------------------------|
| Make | 0: Make<10, 1:10<=Make<18.1 | Original distribution of Make: 0:0.958,1:0.042 Changed distribution of Make: 0: 0.15, 1:0.85 | 0: 0.561, 1: 0.439 |
| SEX | 0: Female, 1: Male | Original distribution of SEX: 0:0.08,1:0.91 Changed distribution of SEX: 0: 0.3, 1:0.7 | 0: 0.632, 1: 0.368 |

*p.d = Probability distribution

Following assumptions are made while generating the counterfactuals for this dataset

- If counterfactuals generated for the given instance on the feature APH is less than that in original dataset ( $X_{APH}^{CF} \leq X_{APH}$ ), then it should be ignored as APH for the selected instance cannot decrease and thus produce invalid counterfactuals.

- SEX generated by the counterfactuals should also be the same as the selected instance.

Table 9 presents that for the 3rd test instance of Insurance fraud dataset, the target variable is non-fraud insurance obtained using Random forest as the black box model. 4 counterfactuals generated using DiCE. The counterfactual values of the features are shown in the highlighted rows. It has been noted that the ACC has been changed from 3 to 1 for all generated counterfactuals. In order to change the target variable to fraudulent insurance, other features' values may also need to be changed in combination with the ACC. It appears that changing the car model from type 5 in the original dataset to type 14 and the DPA from 3 to 2 had no effect on the target variable.

Table 9. Counterfactuals generated by DiCE for 3rd test instance, where CFi are the counterfactual.

| Features | Original Values | CF 1 | CF2 | CF3 | CF4 |
|----------|-----------------|------|-----|-----|-----|
| Make | 5 | | | | 14 |
| PT | 6 | 3 | | | |
| VC | 2 | | | | |
| VPrice | 5 | | | 3 | |
| RepNo | 5 | | | | |
| Ded | 400 | 499 | 672 | | |
| DR | 3 | | | | |
| DPA | 3 | | | | 2 |
| DPC | 2 | | 0 | | |
| AT | 0 | | | | 1 |
| NS | 0 | | | | |
| ACC | 3 | 1 | 1 | 1 | 1 |
| Ncars | 0 | | | 2 | |
| BP | 0 | | | | |
| FraudFound | 0 | 1 | 1 | 1 | 0 |

## 6.4. Credit card Fraud detection dataset

The numeric features were converted into discrete form for categorizing based on the mean value of these features, which is presented in Table 10.

Table 10. Mapping of numeric features based on the mean values for Credit card fraud.

| Feature Name | Mapping |
|---|---|
| V1 | 0:  V1<0 , 1:  0<=V1<3 |
| V2 | 0:  V2<0 , 1:  0<=V2<22.10 |
| V3 | 0:  V3<0 , 1:  0<=V3<10 |
| V4 | 0:  V4<0 , 1:  0<=V4<16.9 |
| V5 | 0:  V5<0 , 1:  0<=V5<35 |
| V6 | 0:  V6<0 , 1:  0<=V6<74 |
| V7 | 0:  V7<0 , 1:  0<=V7<121 |
| V8 | 0:  V8<0 , 1:  0<=V8<21 |
| V9 | 0:  V9<0 , 1:  0<=V9<16 |
| V10 | 0:  V10<0 , 1:  0<=V10<24 |
| V11 | 0:  V11<0 , 1:  0<=V11<13 |
| V12 | 0:  V12<0 , 1:  0<=V12<4 |
| V13 | 0:  V13<0 , 1:  0<=V13<8 |
| V14 | 0:  V14<0 , 1:  0<=V14<11 |
| V15 | 0:  V15<0 , 1:  0<=V15<9 |
| V16 | 0:  V16<0 , 1:  0<=V16<18 |
| V17 | 0:  V17<0 , 1:  0<=V17<10 |
| V18 | 0:  V18<0 , 1:  0<=V18<6 |
| V19 | 0:  V19<0 , 1:  0<=V19<6 |
| V20 | 0:  V20<0 , 1:  0<=V20<40 |
| V21 | 0:  V21<0 , 1:  0<=V21<28 |
| V22 | 0:  V22<0 , 1:  0<=V22<11 |
| V23 | 0:  V23<0 , 1:  0<=V23<23 |
| V24 | 0:  V24<0 , 1:  0<=V24<5 |
| V25 | 0:  V25<0 , 1:  0<=V25<8 |
| V26 | 0:  V26<0 , 1:  0<=V26<4 |

| V27 | 0:  V27<0 , 1:  0<=V27<32 |
| V28 | 0:  V28<0 , 1:  0<=V28<34 |
| Amount | 0:  normAmount<0 , 1:  0<=normAmount<103 |

Fig. 5 depicts one of the DAG obtained by using CausalNex, where we wanted to check how applying intervention on any one of the features affects the distribution of target (or Class). Surprisingly, the features such as Amount, V5, V13, V15,V20,V22,V23,V24,V25,V26,V27 and V28 does not seem to show causal effect on target feature.

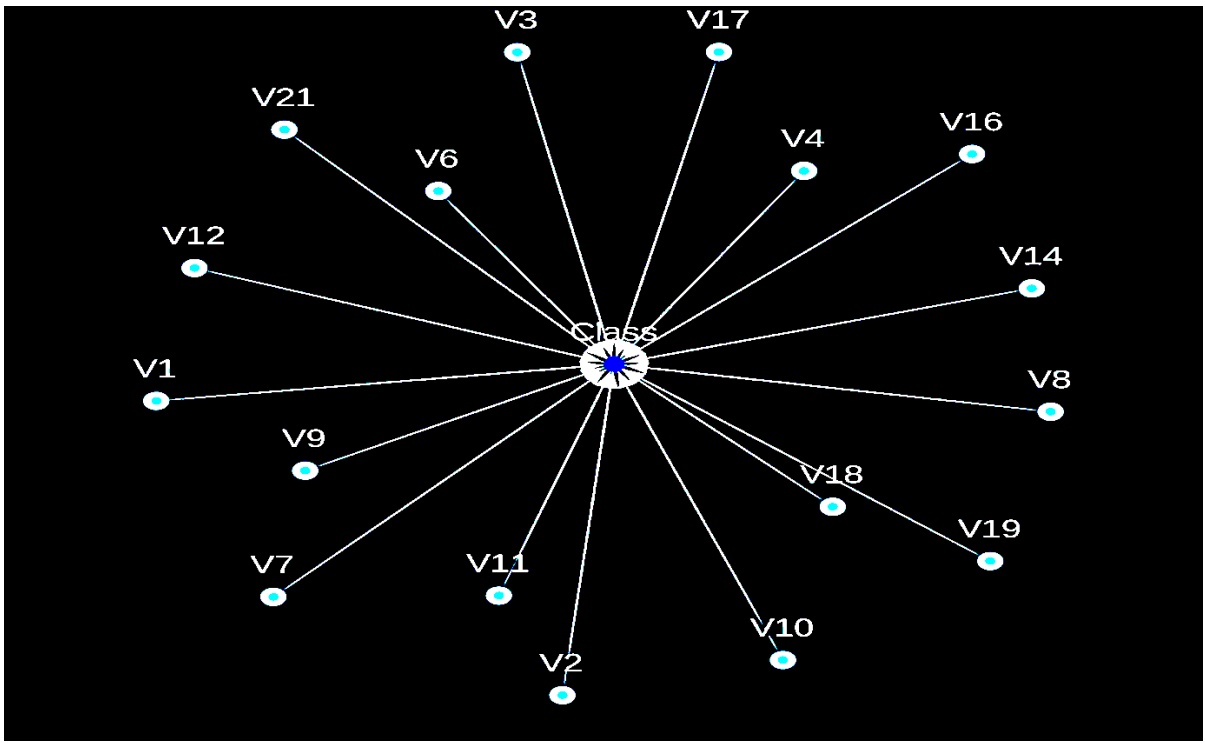

Fig. 5. DAG obtained using CausalNex for Credit card fraud detection dataset.

Table 11 presents the intervention for the feature V1. With respect to this feature, Target variables (Class) are distributed so that 57.1% of them belong to non-fraudulent transactions and the remaining 42.9% are fraudulent transactions. There is a possibility of a decrease in fraudulent transactions from 42.9% to 41.8% if the distribution of V1<0 is increased from 49.6% in the original distribution to 70%.

Table 11. Interventions on features of Credit card fraud dataset

| Features | Category | Intervention | Changed p.d* of target |
|----------|----------|--------------|------------------------|
| V1 | 0: V1<0,<br>1: 0<=V1<3 | Original distribution of feature:<br>0:0.496,1:0.504<br>Changed distribution of feature:<br>0: 0.7, 1:0.3 | 0: 0.582, 1: 0.418 |

*p.d =Probability distribution

Table 12 presents that for the 3rd test instance of Credit card fraud dataset, the target (or class) variable is non-fraudulent transaction obtained using Random forest as the black box model. 2counterfactuals generated using DiCE. The counterfactual values of the features are shown in the highlighted rows. For the 1st generated counterfactual, changing value of V1 from -0.939 to -1.235, V10 from -0.512 to 1.0078 and V19 from 0.3749 to -1.0123 is changing the target variable category from non-fraudulent transaction to fraudulent transactions.

Table 12: Counterfactuals generated by DiCE for Credit card fraud, where CFi are the counterfactual.

| Features | Original Values | CF 1 | CF2 |
|----------|-----------------|------|-----|
| V1 | -0.939 | -1.235 | |
| V4 | -1.1063 | | -1.0322 |
| V10 | -0.512 | 1.0078 | |
| V13 | 0.462 | | 0.2124 |
| V19 | 0.3749 | -1.0123 | |
| normAmount | -0.3531 | | 1.222 |
| Class | 0 | 1 | 1 |

The loan default dataset offers a good set of counterfactuals because just two features need to be changed. However, interventions using CausalNex do not significantly change the target distribution when the distribution of features is changed.

The churn prediction dataset does not provide good counterfactuals, as typically no more than 3 features in each of the generated counterfactuals are intended to be changed, which can be conveniently verified by interventions using CausalNex. Because the distribution of the target is unchanged, although the distribution of features has been changed.

Both insurance fraud and credit card fraud detection shows changes in 3 features, which makes generated counterfactuals nearly good for consideration by the banking and insurance companies.

# 7. Conclusions

In order to incorporate the concept of explainability to solving the ACRM problems, causal inference is adopted here, which is a first-of-its-kind study in ACRM. Four problems in ACRM, namely Loan default dataset, Credit card churn prediction, Credit card fraud detection and Insurance fraud detection dataset are solved using the concepts of causal inference with the help of tools CausalNex and DiCE. Consequently, customers will be provided a set of k valid counterfactuals from banks and insurance companies, allowing for the stronger development of trust between customers and bank staff. Even if it is demonstrated that these features are causally associated in the structural network in the Credit card churn prediction dataset, applying interventions on the same dataset does not result in a change in the target's distribution, demonstrating that they are not related in a causal way. Good quality counterfactuals were produced for Loan default, Insurance fraud, and credit card fraud datasets.

**Appendix A: Tabular description of Classification datasets**

Table A.1. Description of Loan default dataset

| Feature | Description |
|---------|-------------|
| AGE | Age in years |
| BILL_AMT1 | Bill amount statement for September 2005 (NT dollar) |
| BILL_AMT2 | Bill amount statement for August, 2005 (NT dollar) |
| BILL_AMT3 | Bill amount statement for July, 2005 (NT dollar) |
| BILL_AMT4 | Bill amount statement for June, 2005 (NT dollar) |
| BILL_AMT5 | Bill amount statement for May, 2005 (NT dollar) |
| BILL_AMT6 | Bill amount statement for April, 2005 (NT dollar) |
| default.payment.next.month | Target variable with default payment values (1=yes, 0=no) |
| EDUCATION | 1=graduate school, 2=university,3=high school, 4=others, 5=unknown, 6=unknown |
| LIMIT_BAL | Amount of credit granted in NT dollars (including individual and family/supplementary credit). |
| MARRIAGE | Marital status (1=married,2=single, 3=others) |
| PAY_0 | Status of repayment as of September 2005 (-1 = pay on time, 1 = one month's delay, 2 = two months' delay, 8 = eight months' delay, 9 = nine months' delay and above) |
| PAY_2 | Status of repayment as of August 2005 (scale identical to PAY_ 0) |
| PAY_3 | Status of repayment as of July, 2005 (scale same as above) |
| PAY_4 | Status of repayment as of June, 2005 (scale same as above) |
| PAY_5 | Status of repayment as of May, 2005 (scale same as above) |
| PAY_6 | Status of repayment as of April, 2005 (scale same as above) |
| PAY_AMT1 | The amount of the prior payment made in September of 2005 (NT dollar) |
| PAY_AMT2 | Amount of prior payment made in August, 2005 (NT dollar) |
| PAY_AMT3 | Amount of prior payment made in July, 2005 (NT dollar) |
| PAY_AMT4 | Amount of prior payment made in June, 2005 (NT dollar) |
| PAY_AMT5 | Amount of prior payment made in May, 2005 (NT dollar) |
| PAY_AMT6 | Amount of prior payment made in April, 2005 (NT dollar) |
| SEX | Gender (1=male, 2=female) |

Table A.2.Description of Churn prediction dataset.

| Features | Description | Value |
|---|---|---|
| AGE | Customer's age | Positive integer |
| CRED_T | Credit in month T | Positive real number |
| CRED_T-1 | Credit in month T-1 | Positive real number |
| CRED_T-2 | Credit in month T-2 | Positive real number |
| E_CIV | Civilian status | 1– Single, 2 – Married, 3 - Widow, 4 – Divorced |
| INCOME | Customer's income | Positive real number |
| MAR_T | Customer's margin for the company in month T | Real number |
| MAR_T-1 | Customer's margin for the company in month T-1 | Real number |
| MAR_T-2 | Customer's margin for the company in month T-2 | Real number |
| MAR_T-3 | Customer's margin for the company in month T-3 | Real number |
| MAR_T-4 | Customer's margin for the company in month T-4 | Real number |
| MAR_T-5 | Customer's margin for the company in month T-5 | Real number |
| MAR_T-6 | Customer's margin for the company in month T-6 | Real number |
| N_EDUC | Customer's educational level | 1. University student, 2. Medium degree 3. Technical degree, 4. University degree |
| NCC_T | Number of credit cards in month T | Positive integer |
| NCC_T-1 | Number of credit cards in month T-1 | Positive integer |
| NCC_T-2 | Number of credit cards in month T-2 | Positive integer |
| SX | Customers sex | 1 – Male, 0 – Female |
| T_WEB_T | Number of web transactions in month T | Positive integer |
| T_WEB_T-1 | Number of web transactions in month T-1 | Positive integer |
| T_WEB_T-2 | Number of web transactions in month T-2 | Positive integer |
| Target | Target variable | 0 Non churner (nCh), 1 Churner (Ch) |

Table A.3.Description of Insurance fraud dataset

| Feature | Description and Value |
| --- | --- |
| Accident Area | Region of Accident : **Rural, Urban** |
| Address Change Claim | Under 6 months, 1 year, 2 - 3 years, 4 - 8 years, no change |
| Age | From 16 to 80 |
| Age of Policy Holder | 16 - 17, 18 - 20, 21 - 25, 26 - 30, 31 - 35, 36 - 40, 41 - 50, 51 - 65, over 65 |
| Age of Vehicle | 2 years, 3 years, 4 years, 5 years, 6 years, 7 years, more than 7 years, new |
| Agent Type | External, Internal |
| Base Policy | All perils, Collision, Liability |
| Day of week | Sunday to Saturday |
| Day of week claimed | Sunday to Saturday |
| Days Policy Accident | Days left in policy when accident happened: **1 - 7, 15 - 30, 8 - 15, more than 30, none** |
| Days Policy Claims | Days left in policy when claim was filed: **15 - 30, 8 - 15, more than 30, none** |
| Deductible | Amount to be deducted before claim disbursement: **300, 400, 500, 700** |
| Driver Rating | 1, 2, 3, 4 |
| Fault | Policyholder, Third Party |
| Fraud Found | Target variable**: 0, 1 (Fraud Found)** |
| Make | Accura, BMW, Chevrolet, Dodge, Ferrari, Ford, Honda, Jaguar, Lexus, Mazda, Mercedes, Mercury, Nissan, Pontiac, Porsche, Saab, Saturn, Toyota, VW |
| Marital Status | Divorced, Married, Single, Widow |
| Month | January to December |
| Month claimed | January to December |
| Number of Cars | 1, 2 3 - 4 5 - 8, more than 8 |
| Number of Supplements | 1 - 2, 3 - 5, more than 5, none |
| Past Number of Claims | 1,2 - 4, more than 4, none |
| Police Report Filed | Yes, No |
| Policy Number | Policy ID variable for each customers**: 1-15420** |
| Policy Type | Sedan—all perils, Sedan—collision, Sedan—liability, Sport—all perils, Sport—collision, Sport—liability, Utility—all perils, Utility—collision, Utility—liability |
| Rep Number | ID of the Person who processes the claims: **1-16** |
| Sex | Male, Female |
| Vehicle Category | Sedan, Sport, Utility |
| Vehicle Price | (Less than $20,000), ($20,000 - $29,000), ($30,000 - $39,000), ($40,000 - $59,000), ($60,000 - $69,000), (greater than $69,000) |
| Week of Month | 1,2,3,4,5 |
| Week of month claimed | 1, 2, 3, 4, 5 |
| Witness Present | Yes, No |
| Year | 1994, 1995 and 1996 |